\documentclass[10pt,journal,compsoc]{IEEEtran}

\usepackage{graphicx}
\usepackage{hyperref}

\ifCLASSOPTIONcompsoc
  \usepackage[nocompress]{cite}
\else
  \usepackage{cite}
\fi
\ifCLASSINFOpdf
\else
\fi
\begin{document}
%
\title{SlideSpawn: An Automatic Slides Generation System for Research Publications}
%
%
%
%

\author{Keshav Kumar, C. Ravindranath Chowdary
\IEEEcompsocitemizethanks{\IEEEcompsocthanksitem The authors are with the Department of Computer Science and Engineering, Indian Institute of Technology (Banaras Hindu University) Varanasi.\protect\\
E-mail: \{keshav.kumar.cse16, rchowdary.cse\} @iitbhu.ac.in
}}
\IEEEtitleabstractindextext{%
\begin{abstract}
Research papers are well structured documents. They have text, figures, equations, tables etc., to covey their ideas and findings. They are divided into sections like Introduction, Model, Experiments etc., which deal with different aspects of research. Characteristics like these set research papers apart from ordinary documents and allows us to significantly improve their summarization. In this paper, we propose a novel system, SlideSpwan, that takes PDF of a research document as an input and generates a quality presentation providing it's summary in a visual and concise fashion. The system first converts the PDF of the paper to an XML document that has the structural information about various elements. Then a machine learning model, trained on PS5K dataset and Aminer 9.5K Insights dataset (that we introduce), is used to predict salience of each sentence in the paper. Sentences for slides are selected using ILP and clustered based on their similarity with each cluster being given a suitable title. Finally a slide is generated by placing any graphical element referenced in the selected sentences next to them. Experiments on a test set of 650 pairs of papers and slides demonstrate that our system generates presentations with better quality.
\end{abstract}

\begin{IEEEkeywords}
Automatic slides generation, Extractive summarization, Neural networks.
\end{IEEEkeywords}}

\maketitle

\IEEEdisplaynontitleabstractindextext

%
\IEEEpeerreviewmaketitle

\ifCLASSOPTIONcompsoc
\IEEEraisesectionheading{\section{Introduction}\label{sec:introduction}}
\else
\section{Introduction}

\label{sec:introduction}
\fi

%
%
%
%
\IEEEPARstart{P}{resentation} slides have been one of the most popular tools to aid speeches, lectures, seminars since their advent. They have been widely used in the scientific research communities across the world. Researchers often make use of the slides to present their work pictorially in conferences. Creating a presentation from a reference paper is a fairly popular and a time consuming task. It involves listing the important points based on one’s understanding of the paper, organising them in a sensible order and placing the figures, equations and tables at appropriate locations. With the advances in OCR, document parsing and extractive summarization, this task can be easily automated. Presentations can be considered to be a union of textual and graphical elements. So the task of automatically generating slides can be divided into 4 sub tasks.
\begin{enumerate}
    \item Parsing the research document to make use of all the element and their positional information.
    \item Selecting the important textual elements.
    \item Selecting the important graphical elements.
    \item Placing these elements at appropriate locations in the presentation.
\end{enumerate}

In our system, we use a tool called GROBID \cite{GROBID} \cite{GROBID-2} that leverages AI to parse PDF documents and convert them into XML capturing all the elements along with their structural organization. This tool has a special focus on research publications which makes it perfect for our first sub task. To select important textual elements, we use extractive summarization, for this we employ a Multi Layer Perceptron, trained on our dataset (Section \ref{section: dataset}), that tries to predict a salience score for each sentences, We then use ILP to select sentences with highest salience score subject to constraints which ensure that the presentation covers of maximum content of the paper, Finally we cluster the selected sentences into groups and give each group a suitable title by finding a noun phrase that occurs in the group and is semantically close to each sentence. To select graphical elements for our slide, we scan each selected sentence for the PPT for any references to a figure, table or equation, upon finding any, we indicate the element's label in the slide next to the sentence. Finally, we place the sentences within a cluster by sorting them with respect to their position in the paper and sort the cluster with respect to the position of the first sentence in the cluster. We use Google Colaboratory for all our text processing and training tasks.
\begin{figure}
    \centering
    \includegraphics[width=2.5in]{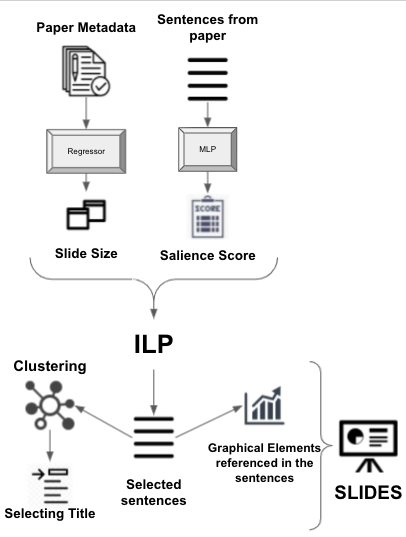}
    \caption{Overview of the SlideSpawn algorithm}
    \label{figure: slideSpawn}
\end{figure}
The main contributions of the paper are the following:
\begin{enumerate}
    \item To the best of our knowledge, we are the first one to use semantic similarity based on sentence vectors to measure a sentence salience score for slides generation.
    \item We propose an end to end system, SlideSpawn, to generate presentations from a reference paper by leveraging it's rich structure.
    \item We release a new dataset, Aminer9.5K Insights, which can be used for automatic summarization tasks and to augment the dataset in automatic presentation generation tasks.
\end{enumerate}

The rest of the paper is organized as follows: Section 2 briefly discusses the related work. Section 3 elaborates the proposed system, SlideSpawn. In section 4 we present some experimental results. Finally, section 5 concludes the paper.

\section{Related work}
There aren't a lot of papers that directly address the problem of Automatic Slide Generation from a Reference Research Paper. Although, the problem of text summarization, which can be seen as a sub-task for our problem, has been worked upon quite extensively. Summarization of research document is different from standard text and summarizing text for placement in a presentation is different from paragraph summarization \cite{slides_differ_from_normal}.
\subsection{Automatic Slides Generation}
The ealiest of the work addressing automatic slides generation can be traced to \cite{Masao}. They propose an algorithm to generate slides from document annotated with GDA tagset. They use abstractive summarization to place sentences in slides. The system proposed in \cite{shibata} generated the slides by itemizing topic and non-topic parts extracted based on syntactic and case analysis. The system proposed in \cite{yoshiaki} converts a tex source to XML, weighs words using Tf-IdF values and assigns a weight to objects like sentence, table, figures etc. while deciding to place them in the slides. The system, SlidesGen \cite{slidesgen}, uses a graph based extractive summarizer QueSTS \cite{Extended_quests} \cite{quests}, to generate section wise slides from a tex source.

The scope of machine learning was very limited due to lack of a dataset in this context. A dataset \cite{slideseer} studying alignment of paper and slide sentences had 10,000 pairs but the dataset remained private. The authors of PPSGen\cite{PPSGen} crawled a dataset of 1,400 pairs following authors profile links listed on aminer, which allowed them to use SVR \cite{SVR} to learn the salience score using a limited set of surface features corresponding to the sentences. They employed ILP to select sentences. In \cite{phrasebased}, authors proposed a phrase-based approach to generate slides, they extract phrases from the give paper and learn their salience and hierarchy, then they greedily place the phrases into the slides. The system proposed in \cite{autoslides} builds on the work done by \cite{PPSGen} using the same dataset. They expand the list of features adding semantics and contextual features to predict sentence salience based on Jaccard similarity. They also use ILP to select sentences for the slides. In \cite{sefidextractive}, authors introduce the PS5K dataset and use a rouge based sentence labelling approach for defining the sentence importance. They use a model based on \cite{sumarunner} and select sentences using ILP.
\subsection{Summarization}
Selecting text for our presentation can be approximated as a task of summarizing the paper. There are two ways to summarize a text \cite{summarization-survey}. First being the extractive approach where we select a subset of the sentences present in the original document. Second being the abstractive approach here, the summary text is generated. In our work, we follow the extractive approach as it gives grammatically and semantically correct sentences \cite{ex-bet-1} \cite{sumarunner} and is time efficient. The task of extractive summarization can be divided into 3 parts \cite{extractive-division}:
\subsubsection{Sentence Representation}
Most of the systems start by finding a way to represent sentences in a vector space that encodes their semantics. Tf-IdF based embedding were popular for a long time. Some later techniques include generating word2vec \cite{word2vec} based embedding \cite{doc2vec}, \cite{doc2vec-survey}, \cite{word2vec-emb-1}. The revolutionary paper by google research on natural language modelling \cite{bert}, that introduced BERT which follows an unsupervised learning approach to learn high level language features, transformed the way we represent sentences. Sentence embedding based on BERT \cite{sentence-bert} have gained more and more traction for a lot of NLP tasks \cite{s-bert-1}, \cite{s-bert-2}, \cite{s-bert-3}. BERT based systems have also been successful in the context of extractive summarization\cite{BERT-sum-1}, \cite{BERT-sum-2}, \cite{BERT-sum-3}. Another development over BERT introduced by Facebook research came with RoBERTa \cite{roberta} which changes some key hyper parameters and training technique to achieve better results on tasks like GLUE \cite{glue}, RACE \cite{RACE}, SQuAD \cite{squad}. These models have a lot of parameters that makes their usage time and resource consuming. Thanks to the distillation \cite{distilBert} technique, we can achieve similar results with higher computation speed. In our work, We use the distil-RoBERTa \cite{sentence-bert} based Sentence Transformer to encode the sentences and measure sentence similarity

\subsubsection{Sentence score}
To decide sentence score, one can use the sentence's similarity with gold-standard sentences. There is a choice of using the average similarity or the maximum similarity, latter giving better results \cite{PPSGen}, \cite{autoslides}. 

\subsubsection{Sentence selection}
The final step is to select the sentences. We already have an importance score, the trivial way to select sentences is to do it greedily. This is effective to some extent but may cause redundancy of information. Another method for sentence selection is solving an integer linear programming (ILP) problem \cite{hongILP}, \cite{gillickILP} which although is an NP hard problem \cite{filatovaNP-hard}, but there are good solvers available that give results in decent time. Using ILP, we can place constraints to reduce redundancy while maximizing the aggregate of sentence score. In our system we use ILP.

\section{SlideSpawn}
In this section we describe the SlideSpawn algorithm. Figure \ref{figure: slideSpawn} shows the overview of the proposed model. We briefly discuss the used dataset, followed by the sentence similarity measure used throughout the pipeline, we then describe our sentence importance assessment model, followed by the presentation generation algorithm.

\subsection{Dataset}
\label{section: dataset}
\begin{figure}[!t]
\centering
\includegraphics[width=2.5in]{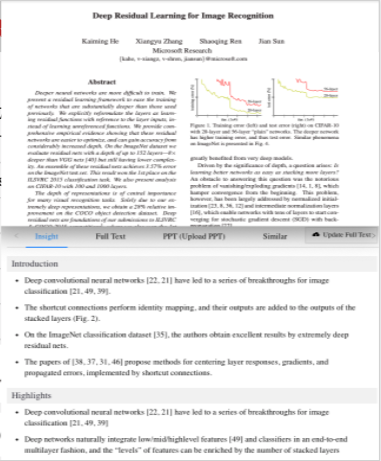}
\caption{Sample pair from Aminer9.5K Insights dataset}
\label{figure:dataset_sample}
\end{figure}

In order to train a machine learning model for our extractive summarization sub task, we use union of two datasets. PS5K \cite{sefidextractive} and Aminer9.5KInsights Dataset, the latter was gathered by us. The PS5K is publically available collection of 5,000 presentation and slides pair compiled from conference proceedings websites. The Aminer9.5K Insights dataset, available here\footnote{\href{https://drive.google.com/file/d/1962dusrdj3yZKyicvykO73id_Z-cplZo/view?usp=sharing}{Aminer9.5K Insights}}, has more than 9,500 research papers along with their sectioned and bulleted summaries taken from aminer\footnote{\url{https://www.aminer.org/}} website. The aminer website, for some of their listed papers, offers an insight section which has sections like Introduction, Highlights, Results etc. and each section has some bullet points that summarize the paper. This is quite similar to the structure of the presentation that SlideSpawn aims to create. This dataset has some of the most cited papers from a wide range of disciplines. A sample pair is displayed in the Figure \ref{figure:dataset_sample}.\\ 
The PPTs provided in the PS5K dataset are in PDF format. Extracting sentences effectively from these files is hard due to their non-uniform structure and presence of many graphical elements. Even with the best tools, we need significant text cleaning to remove unwanted information and some information is lost during extraction. The dataset collected by us avoids this problem altogether as the structure of the insight section is uniform for every paper and we download HTML source of the section, extracting text from which is trivial.\\ 
The final dataset, we use for training our summarization model, has more than 14,500 pairs, with more than 4 million sentences containing over 140,000 unique words.

\subsection{Sentence Similarity}
We use sentence simiarity measures at many points in the system pipeline. Previous systems \cite{PPSGen}\cite{slidesgen}\cite{phrasebased} have adopted cosine similarity over TF-IdF vectors while \cite{autoslides} uses jaccard similarity. Both these methods ignore semantics of the word and focus on exact overlap of the words to measure similarity. Slides are often paraphrased and simplified by authors, and working on a semantic based similarity provides a better alternative while estimating sentence importance for training our extractive summarization model. We use cosine similarity of sentence vector generated by distil-RoBERTa-base \cite{sentence-bert} sentence transformer to calculate sentence similarity in our system. Going forward we refer the vectors generated by distil-RoBERTa-base sentence transformer corresponding to a sentence as simply sentence vectors or sentence embedding and the cosine similarity between two sentence vectors as simply sentence similariy.

\subsection{Sentence Imoprtance Assesment}
\label{subsection: sentence_salience}
\begin{figure}
    \centering
    \includegraphics[width=2.5in]{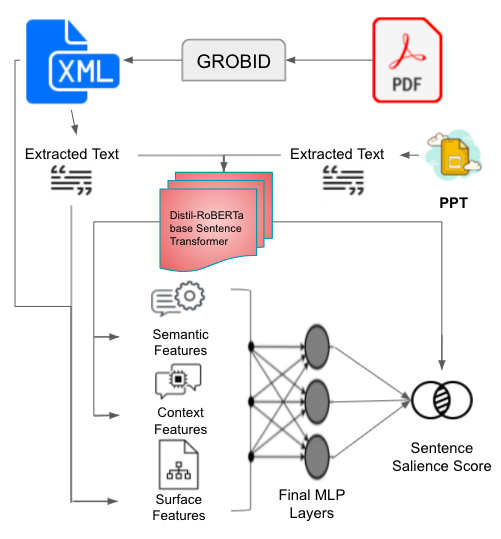}
    \caption{Flow diagram of sentence importance assessment.}
    \label{figure:train_flow}
\end{figure}

The complete flow diagram of sentence importance assessment training can be seen in Figure \ref{figure:train_flow}

\subsubsection{Salience Score}
We define the salience score of a sentence as its maximum similarity with any of the sentences in the presentation or the insight.
\begin{equation}
    S_i\;=\;\max_{j = 0}^{N}\; cosine(EP_i,\; ES_j)
\end{equation}
Where ~$S_i$ is the salience score of the ~$i^{th}$ sentence in the paper, ~$N$ is the total number of sentences in the PPT/Insight, ~$cosine(x, y)$ is the cosine similarity function for two vectors ~$x\; and\; y$, ~$EP_i$ is the sentence vector of the ~$i^{th}$ sentence in the paper and ~$ES_j$ is the sentence vector of the ~$j^{th}$ sentence of the PPT/Insight. This score is the dependent variable for our machine learning model.
\subsubsection{Feature Engineering}
The independent variables, that we calculate corresponding to the sentences, can be broadly classified into following 3 categories:
\begin{enumerate}
    \item Surface features: These capture the syntactic or grammatical attributes of the sentence. These include the number of references to literature, tables, figures, equations present in the sentence, the section(Abstract, Introduction, Background, Model, Results, Conclusion, Acknowledgement) containing the sentence in one-hot format, position of the sentence in the section normalized between ~$(0, 1]$, count of noun and verb phrases in the sentence, number of sub sentences, stopword percentage, length in number of characters and number of tokens, depth of the parse tree of a sentence, average TF and IdF of the words in the sentence.
    \item Semantic features: These capture the meaning of the sentence. These include sentence vectors, sentence similarity with paper title, section title and abstract.
    \item Contextual features: These capture the context of the sentence. We use sentence's similarity with the three sentences that come before it and three sentences that come after it.

\end{enumerate}

\subsubsection{Model and Training}
\begin{figure}
    \centering
    \includegraphics[width=2.5in]{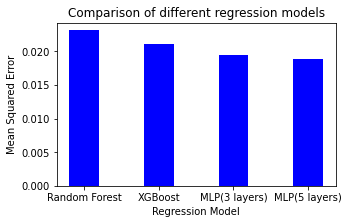}
    \caption{Comparison of different regression models}
    \label{figure: MSE}
\end{figure}

After experimentation with different models and configurations (see Figure \ref{figure: MSE}), we decided to use a multi-layer perceptron for the regression task of predicting the salience score. Our networks has one input layer, three hidden layers and an output neuron, all the layers except the output layer are followed by a sigmoid activation layer. We use Stochastic gradient descent optimizer with learning rate 0.004 and batch size of 64 to train for 50 epochs with the mean squared error as the loss function. 

\subsection{Presentation Generation}

\subsubsection{Sentence Selection}
To select sentences, we use ILP to maximize the objective function given in Equation \ref{equation: objective}, subject to the size constraint (see Equation \ref{equation: size_constraint}) and the similarity constraint that limits the redundancy (see Equation \ref{equation: similarity_constraint}).
\begin{equation}
    \sum_{i = 0}^{N}\;(L_i\; \times \;X_i\; \times \;S_i)
\label{equation: objective}
\end{equation}
Subject to constraints:
\begin{equation}
    \sum_{i=0}^{N}\; (L_i\; \times\; X_i)\;\leq \; Size
\label{equation: size_constraint}
\end{equation}
\begin{equation}
    \frac{\sum_{j=0}^{N}\sum_{i = 0,i\neq j}^{N} (cos(EP_i,\: EP_j)\times X_i\times X_j)}{\sum_{j=0}^{N}\sum_{i = 0,\:i\neq j}^{N}\:(X_i\times X_j)}\leq \Theta
\label{equation: similarity_constraint}
\end{equation}

Where, \\~$N$ : is the number of sentences in the paper.\\
~$L_i$ : is the number of characters in the ~$i^{th}$ sentence.\\
~$X_i$ : is a boolean variable. It is 0 if the ~$i^{th}$ sentence is not selected and 1 if it is selected.\\
~$S_i$ : is the salience score of the ~$i^{th}$ sentence as estimated by our model described in Section \ref{subsection: sentence_salience}\\
~$Size$ : is the number of characters we want in the target presentation.\\
~$cos(x,y)$ : is the cosine similarity function between two vectors ~$x$ and ~$y$.\\
~$EP_i$ : is the sentence vector of the ~$i^{th}$ sentence.\\
~$\Theta$ : is a hyper parameter to be tuned.

To find the target ~$Size$ of the presentation, we use a linear regression model trained on paper's characteristics like its size, the number of sentences, sections, graphical elements, references present in it along with the average size of sentence in terms of tokens and characters labelled with the number of characters in the corresponding presentation in the PS5K dataset.

The hyper-parameter, ~$\Theta$ lets us control the average similarity between the selected sentences. A very low value of ~$\Theta$ compels the model to select sentences with lower salience score and reduces the quality of the summary, while a very high value of ~$\Theta$ makes the selection similar to a greedy selection which increases redundancy in the resulting presentation.

\subsubsection{Presentation Organization}
The text structure in a presentation generally has bullet points at different levels. Each slide has one or more first-level bullets and the first level bullets have several second level bullets that expand on a particular topic. The content distribution of presentation according to an epmirical analysis done by \cite{phrasebased} is listed in Table \ref{table: slide_distribution}. The total share of content on first two levels is over 90\%. Thus we decide to organizes the textual content into two level of bullet points. 

\begin{table}[htb]
\centering

\caption{Bullet points statistics}

\begin{tabular}{|c||c|}
\hline
Levels & Share \\
\hline
\hline
Level 1 & 56.2\% \\
\hline
Level 2 & 38.4\% \\
\hline
\end{tabular}\\
\label{table: slide_distribution}
\end{table}

We first cluster the selected sentences based on their semantic cohesion (see  Section \ref{section: clustering}) and place them at the second level of bullets, we order the sentences according to their position in the paper and order the clusters according to the position of their first sentence. We then, find an appropriate title for the cluster and place it at the first level of bullets. Finally we scan the selected sentences in the XML to find any reference to a graphical element, upon finding a reference, we indicate the id of the graphical element to be placed in the slide next to the sentence.
\subsubsection{Sentence Clusters}
\label{section: clustering}
To cluster the sentences, we make a complete weighted graph with the selected sentences as the nodes and their sentence similarity as the edge weights. We then, binary search for a threshold for the edge weights such that when we remove the edges weighing less than the threshold value, the number of strongly connected components we are left in the graph is closest to ~$N/3$, where ~$N$ is the number of selected sentences. This gives us cluster sizes generally ranging from 1 to 5. The clusters we obtain contain sentences discussing similar topics given in the research paper and can be placed together.

For selecting the title, we consider all the noun phrases present in the cluster as our candidates, and choose the one with highest phrase score defined by the Equation \ref{equation: phrase_sim}.
\begin{equation}
    PS_i = \frac{\sum_{j = 0}^{N} cos(ENP_i,\; EP_j)}{N}
    \label{equation: phrase_sim}
\end{equation}
where, ~$PS_i$ is the phrase score for the ~$i^{th}$ phrase in the cluster, ~$N$ is the number of sentences in a cluster, ~$cos(x, y)$ is the cosine similarity between vectors ~$x, y$, ~$ENP_i$ is the sentence vector for the ~$i^{th}$ phrase, ~$EP_i$ is the sentence vector for the $i^{th}$ sentence in the cluster. 
\section{Results}
Humans evaluate a presentation subjectively which makes devising a quantitative measure for the quality of a presentation a hard task. Since summarization is a key part of our system, we present our evaluations and comparisons using metrics used for the summarization tasks. Most recent summarization works evaluate their models using ROUGE scores\cite{rouge}\cite{rouge-code}. For our comparison, we implement two baseline methods, PPSGen \cite{PPSGen} and Phrase-Based Presentation Slides Generation \cite{phrasebased}.

We randomly separate 650 pairs of paper and PPT/Insight from the training and validation datasets for evaluations and comparison. For each system, we limit the size of presentation for each method to 20\% of the size of the research paper. We use ROUGE-1, ROUGE-2, ROUGE-SU4 F1 scores for comparison. For our model, we select ~$\Theta = 0.55$ as that gives the best results in terms of listed metrics.

\begin{table}[htb]
\centering
\caption{ROUGE Score comparisons.}
\begin{tabular}{||c | c c c||} 
 \hline
Algorithm & Rouge 1 & Rouge 2 & Rouge SU4 \\ [0.5ex] 
 \hline\hline
 PhraseBased & 14.42 & 4.18 & 6.23\\
 \hline
 PPSGen & 34.83 & 22.38 & 26.27\\ 
 \hline
 \textbf{SlideSpawn} & \textbf{38.71} & \textbf{27.84} & \textbf{31.65}\\
 \hline
\end{tabular}
\label{table: results}
\end{table}
The ROUGE-1, ROUGE-2, ROUGE-SU4 F1 scores for each of the model are listed in the Table \ref{table: results}. Our model performs better than existing methods in terms of these metrics by at least four percentage.

\section{Conclusion and Future Work}
This paper proposes a novel system called SlideSpawn to generate presentation slides from research publications. We train a sentence salience assessment model based on Multi-Layer Perceptron and use the ILP method to extract sentences, then place similar sentences together under suitable headings. Experimental results show that our method can generate much better slides than existing methods. We also provide a new dataset, Aminer9.5K Insights, that can be used for automatic summarization and automatic slides generation tasks.

In future, the system can be further improved by: 
\begin{enumerate}
    \item Gathering more presentation and slides pairs, 
    \item Improving PDF to XML conversion for both paper and PPT files,
    \item Selecting textual elements by using abstractive summarization techniques as done by humans.
    \item Using learning to quantify importance of graphical elements.
\end{enumerate}


%





\ifCLASSOPTIONcaptionsoff
  \newpage
\fi

\bibliographystyle{IEEEtran}
\bibliography{bib}

\begin{thebibliography}{10}
\providecommand{\url}[1]{#1}
\csname url@samestyle\endcsname
\providecommand{\newblock}{\relax}
\providecommand{\bibinfo}[2]{#2}
\providecommand{\BIBentrySTDinterwordspacing}{\spaceskip=0pt\relax}
\providecommand{\BIBentryALTinterwordstretchfactor}{4}
\providecommand{\BIBentryALTinterwordspacing}{\spaceskip=\fontdimen2\font plus
\BIBentryALTinterwordstretchfactor\fontdimen3\font minus
  \fontdimen4\font\relax}
\providecommand{\BIBforeignlanguage}[2]{{%
\expandafter\ifx\csname l@#1\endcsname\relax
\typeout{** WARNING: IEEEtran.bst: No hyphenation pattern has been}%
\typeout{** loaded for the language `#1'. Using the pattern for}%
\typeout{** the default language instead.}%
\else
\language=\csname l@#1\endcsname
\fi
#2}}
\providecommand{\BIBdecl}{\relax}
\BIBdecl

\bibitem{GROBID}
P.~Lopez, ``{GROBID:} combining automatic bibliographic data recognition and
  term extraction for scholarship publications,'' in \emph{{ECDL}}, ser.
  Lecture Notes in Computer Science, vol. 5714.\hskip 1em plus 0.5em minus
  0.4em\relax Springer, 2009, pp. 473--474.

\bibitem{GROBID-2}
P.~Lopez and L.~Romary, ``{GROBID} - information extraction from scientific
  publications,'' \emph{{ERCIM} News}, vol. 2015, no. 100, 2015.

\bibitem{slides_differ_from_normal}
W.~Xiao and G.~Carenini, ``Extractive summarization of long documents by
  combining global and local context,'' in \emph{{EMNLP/IJCNLP} {(1)}}.\hskip
  1em plus 0.5em minus 0.4em\relax Association for Computational Linguistics,
  2019, pp. 3009--3019.

\bibitem{Masao}
M.~Utiyama and K.~Hasida, ``Automatic slide presentation from semantically
  annotated documents,'' in \emph{COREF@ACL}.\hskip 1em plus 0.5em minus
  0.4em\relax Association for Computational Linguistics, 1999.

\bibitem{shibata}
T.~Shibata and S.~Kurohashi, ``Automatic slide generation based on discourse
  structure analysis,'' in \emph{{IJCNLP}}, ser. Lecture Notes in Computer
  Science, vol. 3651.\hskip 1em plus 0.5em minus 0.4em\relax Springer, 2005,
  pp. 754--766.

\bibitem{yoshiaki}
Y.~Yasumura, M.~Takeichi, and K.~Nitta, ``\BIBforeignlanguage{English}{A
  support system for making presentation slides},''
  \emph{\BIBforeignlanguage{English}{Transactions of the Japanese Society for
  Artificial Intelligence}}, vol.~18, no.~4, pp. 212--220, Dec. 2003.

\bibitem{slidesgen}
\BIBentryALTinterwordspacing
M.~Sravanthi, C.~R. Chowdary, and P.~S. Kumar, ``Slidesgen: Automatic
  generation of presentation slides for a technical paper using
  summarization,'' in \emph{Proceedings of the Twenty-Second International
  Florida Artificial Intelligence Research Society Conference, May 19-21, 2009,
  Sanibel Island, Florida, {USA}}.\hskip 1em plus 0.5em minus 0.4em\relax
  {AAAI} Press, 2009, pp. 284--289. [Online]. Available:
  \url{http://aaai.org/ocs/index.php/FLAIRS/2009/paper/view/22}
\BIBentrySTDinterwordspacing

\bibitem{Extended_quests}
C.~R. Chowdary, M.~Sravanthi, and P.~S. Kumar, ``A system for query specific
  coherent text multi-document summarization,'' \emph{Int. J. Artif. Intell.
  Tools}, vol.~19, no.~5, pp. 597--626, 2010.

\bibitem{quests}
\BIBentryALTinterwordspacing
M.~Sravanthi, C.~R. Chowdary, and P.~S. Kumar, ``Quests: A query specific text
  summarization system,'' in \emph{Proceedings of the Twenty-First
  International Florida Artificial Intelligence Research Society Conference,
  May 15-17, 2008, Coconut Grove, Florida, {USA}}.\hskip 1em plus 0.5em minus
  0.4em\relax {AAAI} Press, 2008, pp. 219--223. [Online]. Available:
  \url{http://www.aaai.org/Library/FLAIRS/2008/flairs08-054.php}
\BIBentrySTDinterwordspacing

\bibitem{slideseer}
M.~Kan, ``Slideseer: a digital library of aligned document and presentation
  pairs,'' in \emph{{JCDL}}.\hskip 1em plus 0.5em minus 0.4em\relax {ACM},
  2007, pp. 81--90.

\bibitem{PPSGen}
Y.~Hu and X.~Wan, ``Ppsgen: Learning-based presentation slides generation for
  academic papers,'' \emph{{IEEE} Trans. Knowl. Data Eng.}, vol.~27, no.~4, pp.
  1085--1097, 2015.

\bibitem{SVR}
V.~Vapnik, \emph{The nature of statistical learning theory}.\hskip 1em plus
  0.5em minus 0.4em\relax Springer science \& business media, 2013.

\bibitem{phrasebased}
S.~Wang, X.~Wan, and S.~Du, ``Phrase-based presentation slides generation for
  academic papers,'' in \emph{{AAAI}}.\hskip 1em plus 0.5em minus 0.4em\relax
  {AAAI} Press, 2017, pp. 196--202.

\bibitem{autoslides}
A.~Sefid, J.~Wu, P.~Mitra, and C.~L. Giles, ``Automatic slide generation for
  scientific papers,'' in \emph{SciKnow@K-CAP}, ser. {CEUR} Workshop
  Proceedings, vol. 2526.\hskip 1em plus 0.5em minus 0.4em\relax CEUR-WS.org,
  2019, pp. 11--16.

\bibitem{sefidextractive}
\BIBentryALTinterwordspacing
A.~Sefid, P.~Mitra, J.~Wu, and C.~L. Giles, ``Extractive research slide
  generation using windowed labeling ranking,'' in \emph{Proceedings of the
  Second Workshop on Scholarly Document Processing}.\hskip 1em plus 0.5em minus
  0.4em\relax Online: Association for Computational Linguistics, Jun. 2021, pp.
  91--96. [Online]. Available:
  \url{https://www.aclweb.org/anthology/2021.sdp-1.11}
\BIBentrySTDinterwordspacing

\bibitem{sumarunner}
R.~Nallapati, F.~Zhai, and B.~Zhou, ``Summarunner: {A} recurrent neural network
  based sequence model for extractive summarization of documents,'' in
  \emph{{AAAI}}.\hskip 1em plus 0.5em minus 0.4em\relax {AAAI} Press, 2017, pp.
  3075--3081.

\bibitem{summarization-survey}
W.~S. El{-}Kassas, C.~R. Salama, A.~A. Rafea, and H.~K. Mohamed, ``Automatic
  text summarization: {A} comprehensive survey,'' \emph{Expert Syst. Appl.},
  vol. 165, p. 113679, 2021.

\bibitem{ex-bet-1}
Y.~Dong, Y.~Shen, E.~Crawford, H.~van Hoof, and J.~C.~K. Cheung, ``Banditsum:
  Extractive summarization as a contextual bandit,'' \emph{CoRR}, vol.
  abs/1809.09672, 2018.

\bibitem{extractive-division}
A.~Nenkova and K.~R. McKeown, ``A survey of text summarization techniques,'' in
  \emph{Mining Text Data}.\hskip 1em plus 0.5em minus 0.4em\relax Springer,
  2012, pp. 43--76.

\bibitem{word2vec}
T.~Mikolov, K.~Chen, G.~Corrado, and J.~Dean, ``Efficient estimation of word
  representations in vector space,'' in \emph{{ICLR} (Workshop Poster)}, 2013.

\bibitem{doc2vec}
Q.~V. Le and T.~Mikolov, ``Distributed representations of sentences and
  documents,'' in \emph{{ICML}}, ser. {JMLR} Workshop and Conference
  Proceedings, vol.~32.\hskip 1em plus 0.5em minus 0.4em\relax JMLR.org, 2014,
  pp. 1188--1196.

\bibitem{doc2vec-survey}
J.~H. Lau and T.~Baldwin, ``An empirical evaluation of doc2vec with practical
  insights into document embedding generation,'' in \emph{Rep4NLP@ACL}.\hskip
  1em plus 0.5em minus 0.4em\relax Association for Computational Linguistics,
  2016, pp. 78--86.

\bibitem{word2vec-emb-1}
L.~Wu, I.~E. Yen, K.~Xu, F.~Xu, A.~Balakrishnan, P.~Chen, P.~Ravikumar, and
  M.~J. Witbrock, ``Word mover's embedding: From word2vec to document
  embedding,'' in \emph{{EMNLP}}.\hskip 1em plus 0.5em minus 0.4em\relax
  Association for Computational Linguistics, 2018, pp. 4524--4534.

\bibitem{bert}
J.~Devlin, M.~Chang, K.~Lee, and K.~Toutanova, ``{BERT:} pre-training of deep
  bidirectional transformers for language understanding,'' in \emph{{NAACL-HLT}
  {(1)}}.\hskip 1em plus 0.5em minus 0.4em\relax Association for Computational
  Linguistics, 2019, pp. 4171--4186.

\bibitem{sentence-bert}
N.~Reimers and I.~Gurevych, ``Sentence-bert: Sentence embeddings using siamese
  bert-networks,'' in \emph{{EMNLP/IJCNLP} {(1)}}.\hskip 1em plus 0.5em minus
  0.4em\relax Association for Computational Linguistics, 2019, pp. 3980--3990.

\bibitem{s-bert-1}
B.~Wang and C.~J. Kuo, ``{SBERT-WK:} {A} sentence embedding method by
  dissecting bert-based word models,'' \emph{{IEEE} {ACM} Trans. Audio Speech
  Lang. Process.}, vol.~28, pp. 2146--2157, 2020.

\bibitem{s-bert-2}
N.~Thakur, N.~Reimers, J.~Daxenberger, and I.~Gurevych, ``Augmented {SBERT:}
  data augmentation method for improving bi-encoders for pairwise sentence
  scoring tasks,'' \emph{CoRR}, vol. abs/2010.08240, 2020.

\bibitem{s-bert-3}
X.~Cheng, ``Dual-view distilled {BERT} for sentence embedding,'' \emph{CoRR},
  vol. abs/2104.08675, 2021.

\bibitem{BERT-sum-1}
Y.~Liu, ``Fine-tune {BERT} for extractive summarization,'' \emph{CoRR}, vol.
  abs/1903.10318, 2019.

\bibitem{BERT-sum-2}
D.~Miller, ``Leveraging {BERT} for extractive text summarization on lectures,''
  \emph{CoRR}, vol. abs/1906.04165, 2019.

\bibitem{BERT-sum-3}
A.~Srikanth, A.~S. Umasankar, S.~Thanu, and S.~J. Nirmala, ``Extractive text
  summarization using dynamic clustering and co-reference on {BERT},'' in
  \emph{{ICCCS}}.\hskip 1em plus 0.5em minus 0.4em\relax {IEEE}, 2020, pp.
  1--5.

\bibitem{roberta}
Y.~Liu, M.~Ott, N.~Goyal, J.~Du, M.~Joshi, D.~Chen, O.~Levy, M.~Lewis,
  L.~Zettlemoyer, and V.~Stoyanov, ``Roberta: {A} robustly optimized {BERT}
  pretraining approach,'' \emph{CoRR}, vol. abs/1907.11692, 2019.

\bibitem{glue}
A.~Wang, A.~Singh, J.~Michael, F.~Hill, O.~Levy, and S.~R. Bowman, ``{GLUE:}
  {A} multi-task benchmark and analysis platform for natural language
  understanding,'' in \emph{{ICLR} (Poster)}.\hskip 1em plus 0.5em minus
  0.4em\relax OpenReview.net, 2019.

\bibitem{RACE}
G.~Lai, Q.~Xie, H.~Liu, Y.~Yang, and E.~H. Hovy, ``{RACE:} large-scale reading
  comprehension dataset from examinations,'' in \emph{{EMNLP}}.\hskip 1em plus
  0.5em minus 0.4em\relax Association for Computational Linguistics, 2017, pp.
  785--794.

\bibitem{squad}
P.~Rajpurkar, J.~Zhang, K.~Lopyrev, and P.~Liang, ``Squad: 100, 000+ questions
  for machine comprehension of text,'' in \emph{{EMNLP}}.\hskip 1em plus 0.5em
  minus 0.4em\relax The Association for Computational Linguistics, 2016, pp.
  2383--2392.

\bibitem{distilBert}
V.~Sanh, L.~Debut, J.~Chaumond, and T.~Wolf, ``Distilbert, a distilled version
  of {BERT:} smaller, faster, cheaper and lighter,'' \emph{CoRR}, vol.
  abs/1910.01108, 2019.

\bibitem{hongILP}
K.~Hong and A.~Nenkova, ``Improving the estimation of word importance for news
  multi-document summarization,'' in \emph{{EACL}}.\hskip 1em plus 0.5em minus
  0.4em\relax The Association for Computer Linguistics, 2014, pp. 712--721.

\bibitem{gillickILP}
D.~Gillick, B.~Favre, and D.~Hakkani{-}T{\"{u}}r, ``The {ICSI} summarization
  system at {TAC} 2008,'' in \emph{{TAC}}.\hskip 1em plus 0.5em minus
  0.4em\relax {NIST}, 2008.

\bibitem{filatovaNP-hard}
E.~Filatova and V.~Hatzivassiloglou, ``A formal model for information selection
  in multi-sentence text extraction,'' in \emph{{COLING}}, 2004.

\bibitem{rouge}
\BIBentryALTinterwordspacing
C.-Y. Lin and E.~Hovy, ``Automatic evaluation of summaries using n-gram
  co-occurrence statistics,'' in \emph{Proceedings of the 2003 Conference of
  the North American Chapter of the Association for Computational Linguistics
  on Human Language Technology - Volume 1}, ser. NAACL '03.\hskip 1em plus
  0.5em minus 0.4em\relax USA: Association for Computational Linguistics, 2003,
  p. 71–78. [Online]. Available:
  \url{https://doi.org/10.3115/1073445.1073465}
\BIBentrySTDinterwordspacing

\bibitem{rouge-code}
\BIBentryALTinterwordspacing
C.-Y. Lin, ``{ROUGE}: A package for automatic evaluation of summaries,'' in
  \emph{Text Summarization Branches Out}.\hskip 1em plus 0.5em minus
  0.4em\relax Barcelona, Spain: Association for Computational Linguistics, Jul.
  2004, pp. 74--81. [Online]. Available:
  \url{https://www.aclweb.org/anthology/W04-1013}
\BIBentrySTDinterwordspacing

\end{thebibliography}




\end{document}